%% file: main.tex
\documentclass[runningheads]{llncs}
\usepackage[T1]{fontenc}
\usepackage{graphicx}
\usepackage[hidelinks]{hyperref}
\usepackage{multirow}
\usepackage{amsmath}
\usepackage{amsfonts} 
\usepackage{booktabs}
\usepackage{adjustbox}
\usepackage{tabularx,ragged2e}
\usepackage{amsfonts}       % blackboard math symbols
\usepackage{nicefrac}       % compact symbols for 1/2, etc.
\usepackage{microtype}      % microtypography
\usepackage{xcolor}         % colors
\usepackage{todonotes}
\usepackage{placeins}
\usepackage{notoccite}
\usepackage{orcidlink}
\usepackage{comment}
\usepackage{subcaption}
\usepackage{hyperref}

\usepackage{color}

\begin{document}
\title{RAISE: A Unified Framework for Responsible AI Scoring and Evaluation}
\author{Loc Phuc Truong Nguyen\inst{1}\orcidlink{0009-0003-4254-0750} \and
Hung Thanh Do\inst{1}\orcidlink{0009-0005-2025-5080}}
%
% \authorrunning{L.P.T. Nguyen and H.T. Do}
\institute{Friedrich-Alexander-Universität Erlangen-Nürnberg, 91054 Erlangen, Germany 
\email{\{loc.pt.nguyen,hung.t.do\}@fau.de}}
\maketitle              % typeset the header of the contribution
\begin{abstract}
% The abstract should briefly summarize the contents of the paper in
As AI systems enter high-stakes domains, evaluation must extend beyond predictive accuracy to include explainability, fairness, robustness, and sustainability. We introduce RAISE (Responsible AI Scoring and Evaluation), a unified framework that quantifies model performance across these four dimensions and aggregates them into a single, holistic Responsibility Score. We evaluated three deep learning models: a Multilayer Perceptron (MLP), a Tabular ResNet, and a Feature Tokenizer Transformer, on structured datasets from finance, healthcare, and socioeconomics. Our findings reveal critical trade-offs: the MLP demonstrated strong sustainability and robustness, the Transformer excelled in explainability and fairness at a very high environmental cost, and the Tabular ResNet offered a balanced profile. These results underscore that no single model dominates across all responsibility criteria, highlighting the necessity of multi-dimensional evaluation for responsible model selection. Our implementation is available at: \url{https://github.com/raise-framework/raise}.

\end{abstract}

\input{sections/1_Intro}

\input{sections/2_LitRev}
\input{sections/3_Framework}

\input{sections/4_Experiment}

\input{sections/5_Discussion}

\section{Conclusion}
We introduce RAISE, a unified framework that quantifies explainability, fairness, robustness, and sustainability in tabular models, translating high-level regulatory principles into actionable evaluation. Using a performance-controlled study across representative architectures, we observe systematic variation in responsibility profiles, confirming that no single model is universally superior: the MLP is robust and efficient, Tabular ResNet is well balanced, and the Feature Tokenizer Transformer achieves the best fairness at a substantial sustainability cost. Hence, responsible AI centers on selecting the architecture whose trade-off profile fits a specific high-stakes context rather than naming a single “best” model. RAISE provides the practical, modular basis for such context-aware selection and regulatory alignment. Future work will extend coverage to classical models, refine normalization for cross-dataset comparability, and conduct usability studies to validate effectiveness in real-world workflows.

\bibliographystyle{splncs04}
\bibliography{references}

\end{document}

%% file: sections/1_Intro.tex
\section{Introduction}
While regulatory frameworks like the EU AI Act \cite{union2021proposal} mandate responsible AI in high stakes domains, they are fundamentally prescriptive, defining what to achieve but not how to quantitatively verify it. This creates a critical implementation gap that is deepened by a fragmented scientific landscape where powerful tools for individual dimensions have matured in isolation. For instance, fairness toolkits like AIF360 \cite{bellamy2019ai} offer rigorous methods to mitigate bias, yet these interventions can introduce unsustainable computational costs. Similarly, explainability methods like SHAP \cite{lundberg2017unified} provide crucial transparency, but this transparency does not resolve underlying fairness issues, as an explanation can faithfully articulate the logic of a biased model. Consequently, practitioners lack the integrated toolkit needed for a holistic, evidence based risk analysis, preventing them from translating responsible AI principles into verifiable practice.

To address the aforementioned issues, we introduce RAISE (Responsible AI Scoring and Evaluation), a unified framework that systematically quantifies model performance across the foundational and often competing dimensions of explainability, fairness, robustness, and sustainability. We focus specifically on models for structured (tabular) data, as this modality underpins automated decision-making in the most regulated sectors like finance and healthcare, where regulatory demands for transparency and fairness are most acute. Our core methodological innovation is a performance-controlled evaluation that normalizes for predictive F1-Score. This rigor allows us to isolate and compare the inherent responsibility profiles of different model architectures, revealing fundamental and consistent trade-offs across canonical deep learning models like Multilayer Perceptrons, Tabular ResNets, and Transformers. Our work provides a reproducible methodology to operationalize responsible AI, translating abstract principles into an actionable instrument for model selection, auditing, and governance.

\begin{comment}
    The remainder of this paper is organized as follows: We first review the current research landscape to motivate our approach. We then detail our proposed framework and its metrics, followed by its experimental application and results. Finally, we discuss the practical implications of our findings and conclude with our contributions and future work.
\end{comment}

%% file: sections/2_LitRev.tex
\section{Background and Related Work}
A comprehensive evaluation of responsible AI necessitates moving beyond single metrics to a multi-dimensional perspective. This section reviews the state-of-the-art across four foundational pillars of responsible AI, highlighting both the progress within each subfield and the critical gaps that emerge when they are considered in concert.

Explainability, the capacity to link model predictions to input features, is a cornerstone of trustworthy AI. While model-agnostic methods like SHAP \cite{lundberg2017unified} are widely adopted for generating these insights, the field has increasingly moved toward quantitative metrics to formalize evaluation, as exemplified by toolkits like Quantus \cite{hedstrom2023quantus}. Complementing the need for transparency is the imperative for fairness, which aims to mitigate systemic biases that can disadvantage protected groups in high-stakes applications. This goal is supported by a mature ecosystem of formal metrics, such as demographic parity and equalized odds, which are implemented in widely-used toolkits like AIF360 \cite{bellamy2019ai} and Fairlearn \cite{weerts2023fairlearn}. The choice of an appropriate fairness metric is highly context-dependent, reflecting different philosophical and legal interpretations of equity, and remains a critical consideration in any practical deployment.

Beyond these human-centric concerns, responsible deployment also depends on a model's operational integrity, which includes both sustainability and robustness. Sustainability in AI addresses the environmental and resource costs of model training and inference, with established metrics like the Lacoste score \cite{lacoste2019quantifying} to quantify this footprint. Although initially focused on large-scale architectures, these sustainability considerations are increasingly relevant for the structured tabular models that dominate regulated industries. Similarly, robustness measures a model's ability to maintain performance against non-ideal conditions, such as distribution shifts and adversarial attacks. Despite the progress from standardized benchmarks like WILDS \cite{koh2021wildsbenchmarkinthewilddistribution} and RobustBench \cite{croce2021robustbench}, their focus has primarily been on domains like computer vision, leaving robustness for structured tabular data comparatively underexplored.

While holistic evaluation frameworks like HELM \cite{liang2022holistic} and COMPL-AI \cite{guldimann2024compl} represent important progress, their design is fundamentally tailored to large-scale language models. As a result, they provide metrics well-suited for auditing but lack the mechanisms to guide practical decision-making on the trade-offs inherent to regulated, tabular data applications. This leaves a clear and unmet need for a framework that translates multi-dimensional auditing into actionable guidance for responsible model selection.

%% file: sections/3_Framework.tex
\section{Proposed Framework}
\begin{figure*}[ht!]
    \centering
    \includegraphics[width=0.8\linewidth]{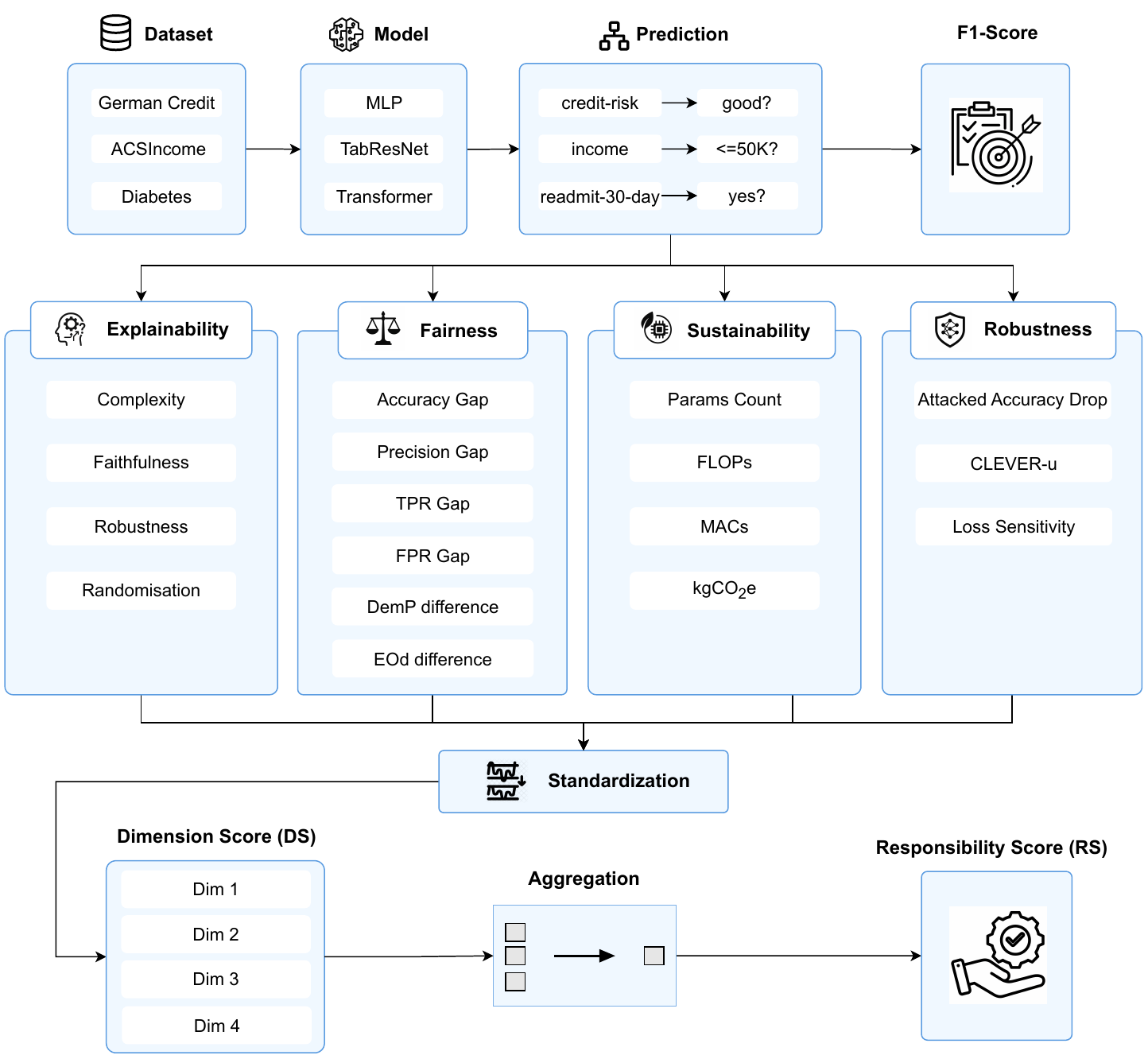}
    \caption{An overview of RAISE.}
    \label{pipeline}
\end{figure*}

RAISE (Responsible AI Scoring and Evaluation) is a unified framework for quantifying model behavior across four core dimensions: explainability, fairness, sustainability, and robustness. As detailed in Figure~\ref{pipeline}, it aggregates established, normalized metrics into a single, interpretable Responsibility Score. Predictive performance is reported separately to enable a direct analysis of the trade-offs between accuracy and responsibility.

\vspace{-1em}

\subsection{Use Cases}
We evaluate our framework on three public, structured datasets representing high-stakes domains: credit risk prediction (finance), diabetes readmission forecasting (healthcare), and income classification (socioeconomics). These tasks were selected because they exemplify real-world scenarios where models are subject to stringent regulatory and ethical scrutiny, making the integrated evaluation of explainability, fairness, robustness, and sustainability not merely beneficial, but essential for responsible deployment.

\subsection{Metric Selection}
We evaluate models using 21 quantitative metrics spanning our four dimensions. This suite, drawn from established literature, provides a comprehensive yet non-exhaustive basis for systematic and reproducible model comparison.

\subsubsection{Explainability}
We assess explainability using a two-stage process. First, we generate model-agnostic feature attributions using SHAP \cite{lundberg2017unified}. Second, we evaluate their quality using eight metrics from the Quantus framework \cite{hedstrom2023quantus}, organized into four categories: explanation robustness, which measures the stability of attributions under input perturbations (\textit{Local Lipschitz Estimate}, \textit{Consistency}); faithfulness, which quantifies their alignment with the model's internal logic (\textit{Faithfulness Correlation}, \textit{Faithfulness Estimate}); randomization, which performs sanity checks against a degraded model (\textit{Model Parameter Randomization Test}, \textit{Random Logit Test}); and complexity, which evaluates the conciseness of the explanation (\textit{Sparseness}, \textit{Complexity}).

\subsubsection{Fairness}
We evaluate fairness by quantifying performance disparities across sensitive subgroups. Our assessment includes measuring the absolute differences in standard classification metrics (Accuracy, Precision, Recall, and False Positive Rate) between groups. We supplement this with two formal group fairness measures from the AIF360 \cite{bellamy2018aifairness360extensible} and Fairlearn \cite{JMLR:v24:23-0389} toolkits: Demographic Parity, which computes the difference in the rate of positive predictions, and Equalized Odds \cite{hardt2016equality}, which measures the disparity in true positive and false positive rates.

\subsubsection{Sustainability}
We assess sustainability by quantifying both environmental impact and computational efficiency. Environmental impact is estimated as carbon emissions (CO2e) using the Lacoste Score \cite{lacoste2019quantifyingcarbonemissionsmachine}, which accounts for hardware power consumption and regional emission factors. Computational efficiency is measured by three standard metrics: the number of parameters, FLOPs, and MACs. To ensure a fair comparison, all sustainability metrics are max-norm scaled across models and datasets.

\subsubsection{Robustness}
We assess model robustness against adversarial perturbations using three metrics implemented with the Adversarial Robustness Toolbox (ART) \cite{art2018}. First, we measure adversarial vulnerability via the \textit{FGSM Accuracy Gap}, which quantifies the drop in test accuracy under attacks generated by the Fast Gradient Sign Method \cite{goodfellow2014explaining}. This is complemented by two attack-independent metrics: the \textit{CLEVER-u Score} \cite{weng2018evaluating}, which estimates the minimum perturbation required to induce misclassification, and \textit{Loss Sensitivity} \cite{arpit2017closer}, which measures the local change in the model's loss in response to input variations.

\subsection{Score Aggregation}
To enable a nuanced comparison, we employ a hierarchical scoring framework. Each raw metric is first normalized to a scale, with lower-is-better values inverted to ensure a score of 1 represents ideal behavior. These are then averaged to produce a Dimension Score (DS) for each of our four pillars. The primary output of our framework is the resulting multi-dimensional responsibility profile, which visualizes the inherent trade-offs across explainability, fairness, robustness, and sustainability. While we also compute a single, aggregated Responsibility Score (RS) for high-level summary, we emphasize the profile as the more informative and actionable tool for nuanced decision-making. Predictive accuracy is reported separately to facilitate this analysis.

%% file: sections/4_Experiment.tex
\section{Experiment and Results}

\subsection{Data and Models}
We evaluate three representative deep learning architectures across three public, high stakes tabular datasets: German Credit \cite{statlog_(german_credit_data)_144} (finance), Diabetes 130-Hospitals \cite{diabetes_130-us_hospitals_for_years_1999-2008_296} (healthcare), and Census Income \cite{adult_2} (socioeconomics). For fairness analysis, we designate gender as the sensitive attribute, reflecting well documented disparities in these domains and ensuring comparability with established benchmarks. While our analysis focuses on this single attribute for clarity, the framework is attribute agnostic and can be readily extended.

To ensure a fair comparison of architectural trade offs, all models were trained to a comparable F1-Score threshold on each dataset. Each experiment was conducted on an 80/20 data split and repeated five times to account for stochastic variability. All models were implemented in PyTorch, with full hyperparameter details provided at: \url{https://github.com/raise-framework/raise}.

\begin{comment}
    Table~\ref{hyperparameters}.

    %\vspace{-2.5em}
    \begin{table}[ht!]
    \centering
    \caption{Hyperparameters used for each model.}
    \renewcommand{\arraystretch}{1.2} % Increase row spacing
    \begin{adjustbox}{width=0.6\textwidth}
    \begin{tabular}{lccc}
    \hline
    \textbf{Hyperparameter} & \textbf{FT-Transformer} & \textbf{TabResNet} & \textbf{MLP} \\
    \hline
    Residual blocks & 3 & 2 & – \\
    Block dimension & 32 & 16 & – \\
    Number of attention heads & 1 & – & – \\
    Attention dropout & 0.2 & – & – \\
    FFN hidden-dim multiplier & 4/3 & – & – \\
    FFN dropout & 0.10 & – & – \\
    Number of hidden layers & 7 & 5 & 1\\
    Hidden dimension & – & 32 & 50 \\
    Dropout & – & 0.2, 0.05 & – \\
    Activation & ReGLU & ReLU & ReLU \\
    Optimizer & Adam & Adam & Adam \\
    Learning rate & 0.001 & 0.001 & 0.001 \\
    Number of epochs & 400 & 400 & 400 \\
    Early stopping patience & 20 & 20 & 20 \\
    \hline
    \end{tabular}
    \end{adjustbox}
    \label{hyperparameters}
    \end{table}
    %\vspace{-1.5em}
\end{comment}

\subsection{Results}
\begin{figure}[ht!]
    \centering
    \begin{subcaptionbox}{German Credit}[.32\linewidth]
        {\includegraphics[width=\linewidth]{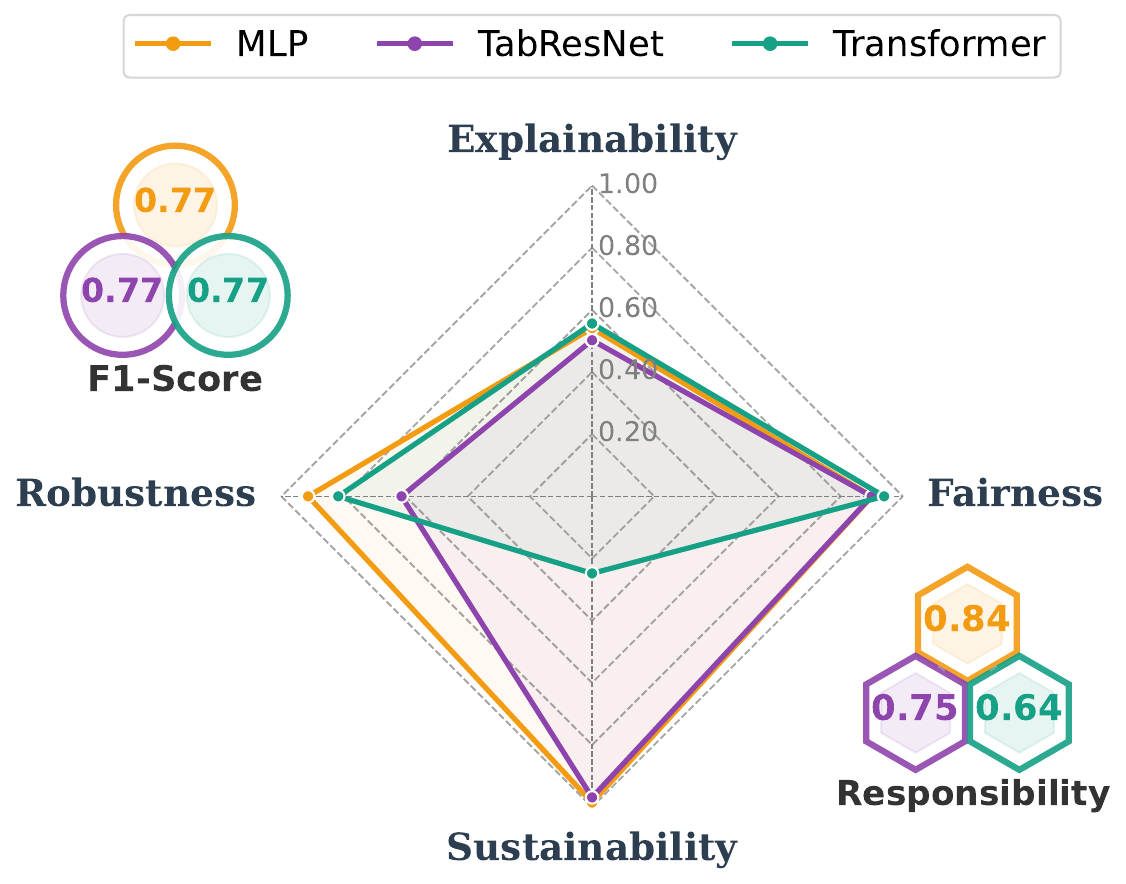}}
    \end{subcaptionbox}
    \begin{subcaptionbox}{ACSIncome}[.32\linewidth]
        {\includegraphics[width=\linewidth]{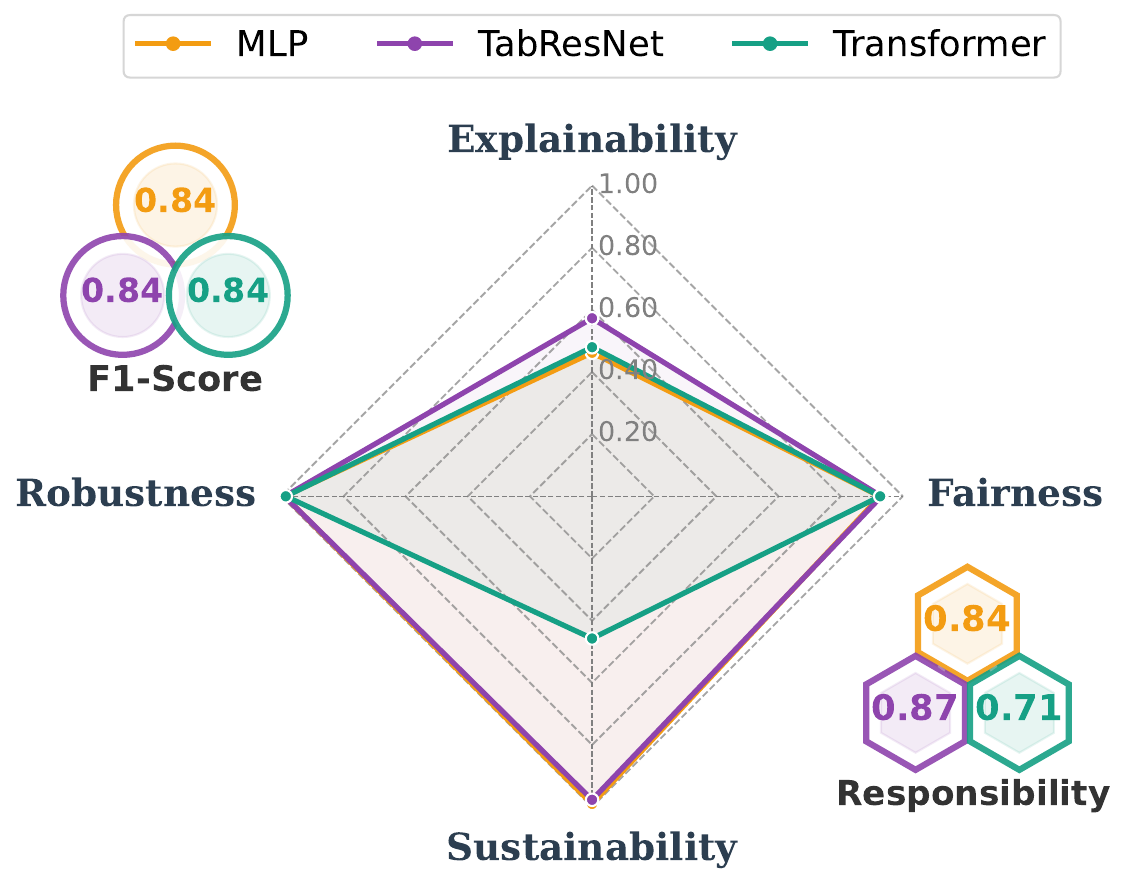}}
    \end{subcaptionbox}
    \begin{subcaptionbox}{Diabetes}[.32\linewidth]
        {\includegraphics[width=\linewidth]{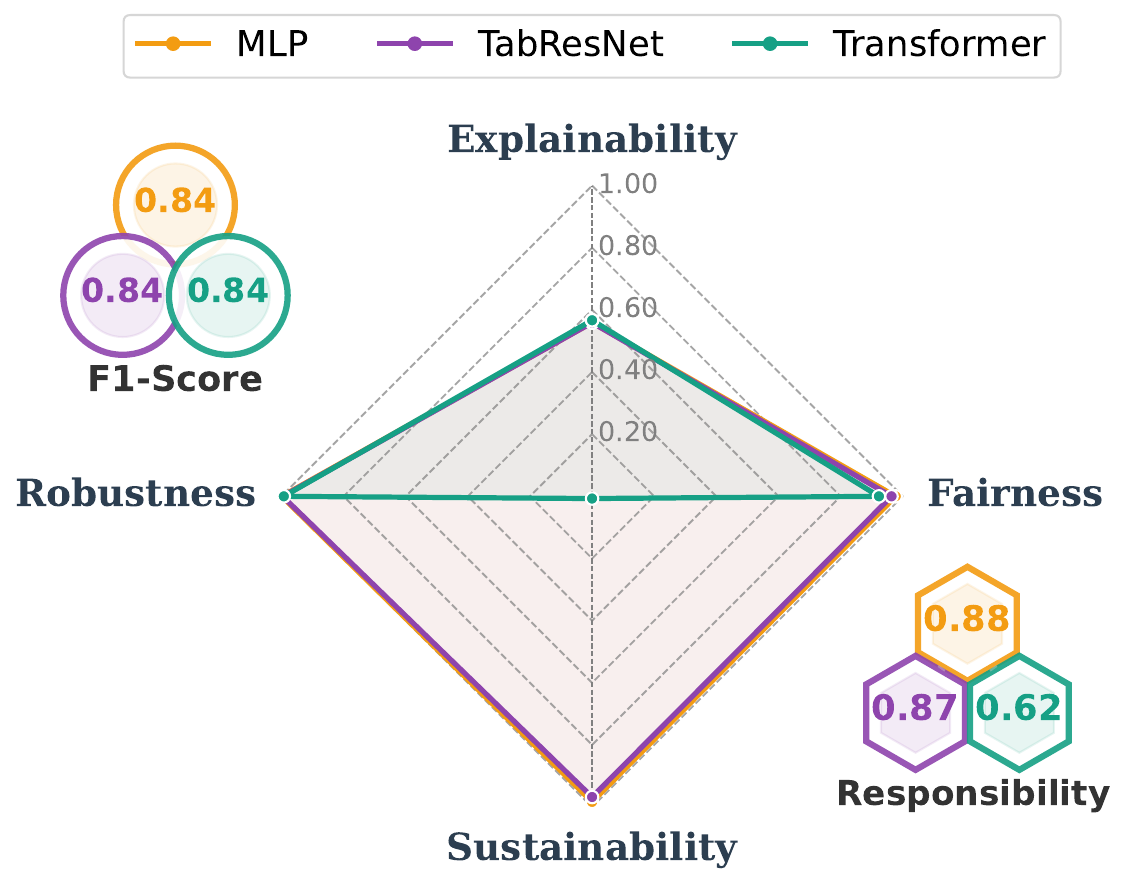}}
    \end{subcaptionbox}
    \caption{Experimental results on three datasets.}
    \label{experiment_results}
\end{figure}

This section reports the evaluation outcomes for all model–dataset pairs across the proposed dimensions. Complete numerical results are presented in Table~\ref{restable}, and Figure~\ref{experiment_results} summarizes the results for each dataset.

%\vspace{-2em}
\begin{table}[ht!]
\caption{Results for all dataset–model pairs under the responsibility framework.}
\label{restable}
\begin{center}
\begin{adjustbox}{width=1\textwidth}
\begin{tabularx}{9.05in}{c *{9}{>{\centering\arraybackslash}X}}
% \begin{tabular}{c *{9}{>{\centering\arraybackslash}p{1.8cm}}} %{c ccc ccc ccc}
\toprule
\textbf{Dataset} & \multicolumn{3}{c}{\textbf{German Credit}} & \multicolumn{3}{c}{\textbf{ACSIncome}} & \multicolumn{3}{c}{\textbf{Diabetes}} \\
\cmidrule[0.5pt]{2-10}
Model   & \multicolumn{1}{c}{    MLP    } & \multicolumn{1}{c}{TabResNet} & \multicolumn{1}{c}{Transformer} 
        & \multicolumn{1}{c}{    MLP    } & \multicolumn{1}{c}{TabResNet} & \multicolumn{1}{c}{Transformer} 
        & \multicolumn{1}{c}{    MLP    } & \multicolumn{1}{c}{TabResNet} & \multicolumn{1}{c}{Transformer} \\
\midrule
% Accuracy
\multicolumn{1}{c|}{\textbf{F1-Score}} 
            & 0.7683 & 0.7715 & \multicolumn{1}{c|}{0.7708} 
            & 0.8362 & 0.8386 & \multicolumn{1}{c|}{0.8444} 
            & 0.8374 & 0.8378 & \multicolumn{1}{c}{0.8379} \\
\midrule
%Responsibility
\multicolumn{1}{c|}{\textbf{Responsibility Score}}
            & 0.8352 & 0.7461 & \multicolumn{1}{c|}{0.6402}
            & 0.8420 & 0.8676 & \multicolumn{1}{c|}{0.7126}
            & 0.8796 & 0.8716 & \multicolumn{1}{c}{0.6222} \\

\midrule
% Explainability
\multicolumn{1}{c|}{\textbf{Explainability Score}} 
            & 0.5412 & 0.5024 & \multicolumn{1}{c|}{0.5562} 
            & 0.4620 & 0.5730 & \multicolumn{1}{c|}{0.4799} 
            & 0.5594 & 0.5589 & \multicolumn{1}{c}{0.5666} \\

\multicolumn{1}{c|}{Complexity} 
            & 0.6697 & 0.7469 & \multicolumn{1}{c|}{0.7476} 
            & 0.6752 & 0.6694 & \multicolumn{1}{c|}{0.6759} 
            & 0.7403 & 0.7523 & \multicolumn{1}{c}{0.7492} \\
\multicolumn{1}{c|}{Faithfulness} 
            & 0.3684 & 0.4011 & \multicolumn{1}{c|}{0.5247} 
            & 0.5501 & 0.6219 & \multicolumn{1}{c|}{0.5710} 
            & 0.6701 & 0.7428 & \multicolumn{1}{c}{0.6372} \\
\multicolumn{1}{c|}{Robustness} 
            & 0.2741 & 0.3288 & \multicolumn{1}{c|}{0.1300} 
            & 0.0527 & 0.1997 & \multicolumn{1}{c|}{0.1267} 
            & 0.0723 & 0.1240 & \multicolumn{1}{c}{0.0685} \\
\multicolumn{1}{c|}{Randomisation} 
            & 0.8524 & 0.5328 & \multicolumn{1}{c|}{0.8225} 
            & 0.5699 & 0.8011 & \multicolumn{1}{c|}{0.5461} 
            & 0.7547 & 0.6166 & \multicolumn{1}{c}{0.8115} \\
\midrule
% Fairness
\multicolumn{1}{c|}{\textbf{Fairness Score}}
                 & 0.9003 & 0.8996    & \multicolumn{1}{c|}{0.9399} 
                 & 0.9264 & 0.9311    & \multicolumn{1}{c|}{0.9271} 
                 & 0.9770 & 0.9636    & \multicolumn{1}{c}{0.9231} \\
                 
\multicolumn{1}{c|}{Accuracy Diff*} 
                & 0.9802 & 0.8889 & \multicolumn{1}{c|}{0.9802}
                & 0.8812 & 0.8868 & \multicolumn{1}{c|}{0.8812}
                & 0.9541 & 0.9562 & \multicolumn{1}{c}{0.9609} \\

\multicolumn{1}{c|}{Precision Diff*} 
                & 0.9544 & 0.8727 & \multicolumn{1}{c|}{0.9033}
                & 0.9256 & 0.9747 & \multicolumn{1}{c|}{0.9886}
                & 0.9643 & 0.9165 & \multicolumn{1}{c}{0.7682} \\

\multicolumn{1}{c|}{TPR Diff*} 
                & 1.0000 & 0.9637 & \multicolumn{1}{c|}{0.9319}
                & 0.9536 & 0.9320 & \multicolumn{1}{c|}{0.8903}
                & 0.9899 & 0.9822 & \multicolumn{1}{c}{0.9647} \\

\multicolumn{1}{c|}{FPR Diff*} 
                & 0.6667 & 0.8730 & \multicolumn{1}{c|}{0.9444}
                & 0.9452 & 0.9308 & \multicolumn{1}{c|}{0.9482}
                & 0.9999 & 0.9996 & \multicolumn{1}{c}{0.9987} \\

\multicolumn{1}{c|}{DemP Diff*} 
                & 0.8929 & 0.9524 & \multicolumn{1}{c|}{0.9841}
                & 0.8603 & 0.8331 & \multicolumn{1}{c|}{0.8575}
                & 0.9972 & 0.9956 & \multicolumn{1}{c}{0.9926} \\

\multicolumn{1}{c|}{EOd Diff*} 
                & 0.6667 & 0.8730 & \multicolumn{1}{c|}{0.9319}
                & 0.9452 & 0.9308 & \multicolumn{1}{c|}{0.8903}
                & 0.9899 & 0.9822 & \multicolumn{1}{c}{0.9647} \\
\midrule
% Sustainability
\multicolumn{1}{c|}{\textbf{Sustainability Score}}
                 & 0.9855 & 0.9689    & \multicolumn{1}{c|}{0.2480} 
                 & 0.9899 & 0.9766    & \multicolumn{1}{c|}{0.4575} 
                 & 0.9833 & 0.9677    & \multicolumn{1}{c}{0.0071} \\

\multicolumn{1}{c|}{Parameters Count*} 
                & 0.9513 & 0.8973 & \multicolumn{1}{c|}{0.0199}
                & 0.9708 & 0.9455 & \multicolumn{1}{c|}{0.0000}
                & 0.9513 & 0.8973 & \multicolumn{1}{c}{0.0286} \\

\multicolumn{1}{c|}{FLOPs*} 
                & 0.9978 & 0.9955 & \multicolumn{1}{c|}{0.0000}
                & 0.9987 & 0.9958 & \multicolumn{1}{c|}{0.4649}
                & 0.9978 & 0.9955 & \multicolumn{1}{c}{0.0000} \\

\multicolumn{1}{c|}{MACs*} 
                & 0.9972 & 0.9943 & \multicolumn{1}{c|}{0.0000}
                & 0.9983 & 0.9946 & \multicolumn{1}{c|}{0.4342}
                & 0.9972 & 0.9943 & \multicolumn{1}{c}{0.0000} \\

\multicolumn{1}{c|}{Normalized kgCO2e*} 
                & 0.9958 & 0.9887 & \multicolumn{1}{c|}{0.9723}
                & 0.9920 & 0.9704 & \multicolumn{1}{c|}{0.9308}
                & 0.9868 & 0.9836 & \multicolumn{1}{c}{0.0000} \\
\midrule
\multicolumn{1}{c|}{\textbf{Robustness Score}} 
            & 0.9139 & 0.6133    & \multicolumn{1}{c|}{0.8168}
            & 0.9898 & 0.9895    & \multicolumn{1}{c|}{0.9858}
            & 0.9988 & 0.9960    & \multicolumn{1}{c}{0.9921} \\
\multicolumn{1}{c|}{Accuracy Gap*} 
            & 0.9500   & 0.9600   & \multicolumn{1}{c|}{0.9900} 
            & 0.9943   & 0.9983   & \multicolumn{1}{c|}{0.9989} 
            & 1.0000   & 1.0000   & \multicolumn{1}{c}{1.0000} \\
\multicolumn{1}{c|}{CLEVER-u} 
            & 0.9965   & 0.8800   & \multicolumn{1}{c|}{0.9195} 
            & 0.9780   & 0.9735   & \multicolumn{1}{c|}{0.9600}
            & 0.9975   & 0.9880   & \multicolumn{1}{c}{0.9765} \\
\multicolumn{1}{c|}{Loss Sensitivity*} 
            & 0.7951 & 0.0000    & \multicolumn{1}{c|}{0.5410}
            & 0.9972 & 0.9968    & \multicolumn{1}{c|}{0.9986}
            & 0.9989 & 0.9990    & \multicolumn{1}{c}{0.9999} \\
\bottomrule
\multicolumn{10}{l}{\textbf{Note}: Metrics marked with an asterisk (\small *) are lower-is-better by definition. Their values have been inverted using $1-\text{raw}$ to ensure consistent scoring direction.} \\
\end{tabularx}
\end{adjustbox}
\end{center}
\end{table}

%\vspace{-3em}
%\vspace{-3em}

Our evaluation shows that key trade-offs are built into each architecture. The Feature Tokenizer Transformer performed very well on nuanced tasks, offering strong explainability and fairness, especially in the difficult low-data setting. However, this advantage came with a significantly high cost in terms of sustainability. In contrast, the simple MLP was relatively robust and energy-efficient but produced quite broad and less faithful explanations. The Tabular ResNet consistently delivered a balanced profile, acting as a reliable middle ground between these two ends and maintaining steady results across conditions.

Importantly, these large differences in responsibility were hidden by the fact that all models reached similar F1 scores. This result shows that predictive accuracy is a weak and often misleading stand-in for a model’s real operational and ethical fitness. It therefore shifts how we think about responsible model selection: the goal is not to identify a single best architecture, but to make a careful choice of the architectural profile whose built-in trade-offs best match the specific ethical and operational needs of the target application.

%% file: sections/5_Discussion.tex
\section{Discussion}

Our work challenges a core assumption in applied AI: that "better" simply means more accurate. For too long, the field's focus on accuracy leaderboards has been a dangerous oversimplification, hiding critical risks in fairness and reliability. The real purpose of RAISE is to provide a more complete picture. It is a tool designed to make the hidden trade-offs visible, creating a clear and defensible record of why a particular model was chosen. This shifts the goal from simply chasing a higher score to engineering a solution that is demonstrably safe and aligned with the values of a specific real-world context.

Building on this shift in objective, RAISE provides the modular and reproducible foundation for evidence-based governance. However, we identify three key directions for future work. First, we will expand the framework to include the classic, non-neural models like boosted trees that are still workhorses in many industries. Second, we will add privacy as a core dimension, measuring how well a model protects sensitive data. Finally, and most importantly, we need to move beyond the lab. We plan to work directly with stakeholders to see how our framework helps them make better, safer decisions in their daily work, ensuring our technical solution becomes a genuinely useful instrument for responsible governance.